Title: Teaching Turn-Taking Skills to Children with Autism using a Parrot-Like Robot

Authors: Pegah Soleiman (1), Hadi Moradi (1 and 2), Maryam Mahmoudi (3), Mohyeddin Teymouri (4) , Hamid Reza Pouretemad (5) ((1) School of ECE, University of Tehran, North Karegar St., Tehran, Iran, (2) Intelligent Systems Research Institute, SKKU, Suwon, South Korea, (3) Department of Psychology, Allameh Tabatabái University, Tehran, Iran, (4)  Department of speech therapy, Iran university of medical sciences, Tehran, Iran (5) Department of Psychology, Shahid Beheshti University, Tehran, Iran)

Robot Assisted Therapy (RAT) is a new paradigm in many therapies such as the therapy of children with autism spectrum disorder. In this paper, we present the use of a parrot-like robot as an assistive tool in turn-taking therapy. The therapy is designed in the form of a card game between a child with autism and his/her therapist or the robot. The intervention was implemented in a single subject study format and the effect sizes for different turn-taking variables are calculated. The results show that, the child-robot interaction had larger effect size than the child-trainer effect size in most of the turn-taking variables. Furthermore, the therapist's point of view on the proposed Robot Assisted Therapy is evaluated using a questionnaire.  The therapist believes that the robot is appealing to children which may ease the therapy process. The therapist suggested to add other functionalities and games to let children with autism to learn more turn taking tasks and better generalize the learned tasks.

**Table of Content**

# 1 Introduction

Children with Autism Spectrum Disorder (ASD), have considerable difficulties in communication and social skills, such as eye contact, verbal communication, imitation, and interacting with other people. They also show restricted and repetitive behaviors, such as fixated interests and hyperactivity to sensory inputs (Bell 1994). The typical approach in dealing with this disorder is human-based intervention in which therapists or parents have to work extensively with these children. Unfortunately, the number of these children is increasing in recent years and the typical intervention approach becomes very hard or impractical. This is the reason behind Robot-Assisted Therapy (RAT) which is used in therapy of children with developmental disorders like ASD (Atherton et al. 2011; Robins et al. 2009). The use of robots in therapy sessions may reduce the load on therapists or parents. Furthermore, the robots may be more effective due to their attractiveness.

Many recent studies have used available robots, such as NAO (Shamsuddin et al. 2012), and investigated their effectiveness in the field of autism therapy. To train social skills such as eye contact and social touch, (Yun et al. 2016) used a mobile humanoid robot iRobi, which performed training scenarios in accordance with Discrete Trial Teaching (DTT) protocol. Their results showed that children were more active and concentrated to the training requests in the clinical trials. Zheng et al. 2014) used a robotic system to learn imitation skills. This system has real-time performance evaluation and feedback and the results show relatively better performance than a human therapist. There are many studies in which NAO, a humanoid robot, has been used for autism therapy, such as training joint attention skills (Anzalone et al. 2014). The results show the effectiveness of using the robot while more studies are needed to bring the robot to therapy floor. (Vanderborght et al. 2012) used a huggable robot, called Probo, that is used as a story teller agent. Probo-oriented therapy showed better improvement of social behavior in children with autism compared to traditional therapy. Keepon is a small creature-like robot (Kozima et al. 2009) that has minimal design of appearance and behavior to help children better understand robot's attention and emotion. The children showed various spontaneous interactive behavior with Keepon.

Based on many studies, such as the ones above, it is believed that robots are more appealing and accepted by children with autism since human-robot interaction does not have complexities involved in human-human interaction. Furthermore, a robot's actions and behaviors can be controlled by therapists/operators to conduct a desired/planned therapy session. This is hard or impossible if done in a group of children. Also, beside time and cost issues, it needs adequate synchronizations if two therapists join to conduct a therapy session such as turn-taking therapy session.

Along this trend, we decided to design a robot for verbal communication therapy. That is why we designed a parrot-like robot that is both attractive and is famous for its ability to speak. Parrots are loved by people since they can verbally interact with them, which is a unique capability among animal pets.

On the other hand, many children with autism miss turn-taking skills, which is important in successful verbal communication. Consequently, we decided to develop therapy protocols, based on our parrot-like robot, to improve this set of skills.

In the first step, we started with verbal turn-taking which is an important skill for success in different social situations, such as in dialogue with other people. Also, it is very time consuming and requires huge dedication from therapists. Consequently, robots can be helpful to reduce the stress and burden on therapists and ease up the therapy process.

The therapy scenario is designed based on a turn-taking game between a child and a partner, which could be a robot, i.e. child-robot therapy, or a therapist, i.e. child-trainer therapy. The game contains three category of things, i.e. fruits, animals, and body parts to be named by the child or his/her partner. Three variables, i.e. turn-taking, turn-telling, and commanding were determined to be evaluated during the therapy sessions. Turn-taking is referred to the behavior in which the child acts upon his/her turn. Turn-telling is referred to the behavior in which the child determines the correct turn, i.e. he/she answers the "whose turn is it?" question. Finally, commanding is referred to the behavior in which the child tells the robot/trainer that is their turn. We determined a group of sub-variables for each of these three variables that are shown in Table 1 and described below:

- **Non-Directed turn taking (ND):** The subject performed turn taking without any help from others.
- **Directed turn taking:** the subject performed turn taking with help from others which can be done verbally, i.e. Verbally Directed (VD) turn takin, or by pointing and physically, i.e. Physically Directed turn taking (PD).
- **Spontaneous Command (SC):** the subject commanded the partner to take his/her turn on his/her own without any help or guidance.
- **Directed Command (DC):** the subject needed guidance and help to correctly tell the partner his/her turn.
- **No Command:** No command initiated by the subject, neither spontaneous nor with guidance, in a session.
- **False Answer (FA):** the subject could not correctly determine the turn at a given instance.
- **Correct Answer (CA):** the subject could correctly determine the person whose turn is at a given instance.
- **Correct Answer Directed (CAD):** the subject needed guidance to correctly determine the turn at a given instance.

For assessing child-robot and child-trainer sessions, quantitative and qualitative approaches can be used. In the quantitative approach, the variables and sub-variables can be analyzed based on their effect sizes. In the qualitative approach, the feedback from trainers, in an interview form, can be collected. Furthermore, the therapy sessions' videos can be evaluated by experts to determine the effectiveness of the child-robot vs. child-trainer therapy.

## 2  Method

### 2.1  The Robot.

Our parrot-like robot, which is called RoboParrot, is based on a toy from Hasbro Toy Company. The mechanical part of the robot contains two motors which enable the robot to move its body, beak, eyes, and wings. In order to give the control of the robot to the operator, its controller board and sensors are replaced. A touch sensor is embedded in its beak to detect the touch of the beak. A program has been developed to control the robot which provides several actions such as random movement or plays sound when touched. The program can be used by the operator to execute a specific action, such as laughing, or a complex task such as face recognition. Also, the robot has two groups of voices: one group includes basic interaction words and turn-taking items, another one includes parrot natural voice and funny laughing to attract children. An operator controls the robot from a computer which could be wired or remotely (Soleiman et al. 2014; Soleiman et al. 2015).

Participants. To better understand the problem under study, purposeful and criterion-based sampling was used (Creswell 2007). To increase the quality of data, individuals were selected using criterion-based subject selection. Thus, a card-based turn-taking test was designed which evaluated basic turn-taking ability. This test was performed on a pool of 28 children with ASD, from which 19 children did not show turn-taking ability. Based on the family commitment and the availability of therapists, three children started the therapy sessions. Unfortunately, due to family reasons, two children were dropped from the study at the beginning of the study and only one child finished the whole therapy session. Based on the GARS score the child is classified in medium autism severity.

### 2.2  Experimental design

A single subject study using cross treatment, cross variables and AB design was used to assess the effectiveness of the current training program. The cross treatment consisted of two distinct treatments, i.e., treatment by a trainer and treatment by the robot. The cross variables included three variables, i.e., the number of successful turn-takings, the number of successful turn-telling, and the number of successful uses of turn command. An AB design contains two phases, i.e. a baseline ("A" phase) with no intervention, and an intervention ("B") phase. Initially, we conducted baseline, and then performed the intervention phase.

Experimental design. Since we had only one subject, a single subject experiment was designed. To eliminate the effect of treatment order, cross treatment was used in which the order of child-robot and child-trainer treatments was changed in each therapy session. Because of three different therapies for the three variables i.e., the number of successful turn-takings, the number of successful turn-telling, and the number of successful use of turn command, multiple-baseline design was used. To get the baselines, an AB design was used in which the baseline, ("A" phase) with no intervention, was performed followed by the intervention phase ("B" phase).

## 2.3 Procedure.

During the subject selection process, 28 children with ASD were tested to see if they accept the robot. The reason behind this was our willing to observe and evaluate the reaction of the children to the presence of robot in the therapy. In this process, three group of cards, with pictures of animals, body parts, and fruits, were provided to the subjects to be named by the subjects and the robot in turn. Obviously, besides selecting the subjects who were willing to play with the robot, the subjects with no ability of correctly naming the cards were considered. Finally, three subjects were selected to participate in the therapy.

All of the therapy sessions were recorded for later evaluation along with an interview form that was designed for therapists.

**Baseline (A).** To determine the initial level of the subjects, multiple-baseline sessions were conducted in which the cards were shown to the subjects and they were asked to name them in turn with the therapist/robot. In the child-trainer scenario, the trainer explained the scenario to the subject. He named a card and then asked the subject to do the same with another card. Then, after finishing this child-trainer scenario the trainer explained the scenario in the child-robot setup to the subject. In this scenario, the robot started the interaction.

The trainer verbally (VD) or physically directed (PD) the subjects when they were unsuccessful in determining the turns. We had 2 baseline sessions for turn-taking variable, 6 sessions for turn-telling variable and, 8 sessions for commanding variable.

**Intervention (B).** Every therapy session lasted, at least, 6 minutes in which 3 to 4 minutes were allocated to child-trainer scenario and 3 to 4 minutes dedicated to child-robot scenario. As it can be seen in Table 2, the turn-taking therapy conducted for 15 sessions, the turn-telling therapy conducted in 11 sessions, and the commanding therapy lasted 9 sessions. The therapy sessions were conducted once or twice a week. Cross treatment protocol was implemented by swapping the starting therapy scenario in each session. That is, if a therapy session started with child-trainer scenario, in the next session the therapy started with child-robot scenario.

**2.4    Tools**

To evaluate the effectiveness of the proposed therapy from therapist point of view, we prepared an open-ended questions interview, with 6 questions.

Analysis method. A mixed method approach consisted of quantitative and qualitative analysis was chosen to understand the effectiveness of the therapy.

2.4.1    Quantitative analysis

For quantitative analysis, we used descriptive findings and inferential results. The frequency ratios for all sessions' data, mean and standard deviation for baselines and intervention sessions were considered as descriptive findings.

We aggregated ND+VD because we wanted to know the times that the child collaborated in the therapy and took turn with his partner without any physical enforcement. Also, we aggregated RA+RAD to discriminate all right answers (even with direction) from false answers. At the end, we aggregated SC+DC to determine all commands that the subject used during the sessions even with direction.

Effect Sizes (ESs) is used as inferential results which assesses differences between groups or variables (Durlak 2009). To calculate effect size, Standardized Mean Difference (SMD), Percentage of data points Exceeding the Median (PEM), and Non-overlap of All Pairs (NAP) were considered in this study. These are briefly explained as follows.

**SMD.** SMD is calculated as the difference between the mean of the baseline phase from the mean of the intervention phase divided by the pooled, or average, of the two groups' standard deviations. An SMD of zero means that the intervention and the baseline conditions had equivalent effects. SMDs greater than zero indicate the degree to which intervention is more efficacious than baseline conditions and SMDs less than zero indicate the degree to which intervention is less efficacious than baseline conditions (Faraone, 2008). (Cohen 1988) offered the following guidelines for interpreting the magnitude of the SMD in the social sciences: small effect, SMD = 0.2; medium effect, SMD = 0.5; and large effect, SMD = 0.8.

**PEM.** PEM shows the percentage of data points using a new approach, i.e. the child-robot therapy in our case, to the median of a typical method, i.e. the child-trainer approach in our study.

**NAP.** Non-overlap of All Pairs (NAP) is another method for evaluating a new therapy method compared to a traditional one. NAP compares the non-overlapped points to the total number of points which is a representation of the data points passing the traditional method's data points.

2.4.2   Qualitative analysis

Since we wanted to evaluate the efficacy of child-robot intervention from therapists' point of view, the interview form was provided and used for qualitative analysis. In addition, to assess which differences were observed between child-trainer interaction and child-robot interaction, all session video recordings were observed and evaluated.

# 3   Results

As mentioned earlier, total of 28 children were tested from which 19 were selected for intervention. However, only three children were available to start the study. Unfortunately, after a few sessions only one child were left and the other two left the sessions due to family relocation or issues.

## 3.1   Quantitative results

SMD: To evaluate the results, we turned all raw data into frequency ratios shown in Tables 2, 3, and Figures. 1, 2. Turn-taking training consisted of two baseline sessions followed by 15 intervention sessions and turn-telling training consisted of six baseline sessions followed by 11 intervention sessions. It should be noted that the subject did not have turn-telling at all. Thus, the actual turn-telling base-line evaluation was performed at the beginning of the 5th and 6th sessions. Finally, the subject did not have self-commanding at all until the commanding therapy started. Furthermore, as it can be seen in Table 2, due to the difficulties that the subject had, in S1, S3, S5, S6, S8, and S9, the directed commanding intervention was not explicitly performed in the child-trainer sessions. Rather than that, the behaviors of the child were observed to check if the effect of observations.

In contrast to the child-trainer therapy, the child was more cooperative in the child-robot therapy. That is why we could perform the baseline evaluation for the commanding therapy in child-robot scenario (Table 3) while it was not possible in child-trainer scenario (Table 2).

Figures 2, 3, and 4 show both the child-robot and the child-trainer therapies with their baselines. For the child-robot therapy the slope of the child progress is ascending for all variables but for the child-trainer one, ND+VD and SC+DC have descending slope. It might be because the child was not in mood and refused to collaborate with the trainer in the last sessions. Even for RA+RAD, in which the child-trainer therapy had ascending slope, the numbers are not a stable one while it converges to 100 for the child-robot therapy. As it can be seen in Figure. 4 the child improvement in commanding for the child-robot therapy has a jump in session 5. Based on the trainer's opinion, it may be because of child's good learning ability in simple tasks. But after session 7 he was not in good mood so it shows a little drop.

Similar to typical therapy sessions, the child was inattentive and distracted in a few sessions. Consequently, his progress had ups and downs. Thus, we need to further test the subject and/or perform it on a bigger pool of subjects.

Table 4. shows the means and standard deviations of the Child-robot and the Child-trainer baseline and therapy sessions for the three aggregated sub-variables.

To compare the child-robot and the child-trainer approaches, three SMDs were calculated based on different baselines (Table 5). The child-trainer baseline is the typical way of assessing the competence of the child in turn-taking. In such a case, the trainer used a set of cards to determine if the child could do turn taking, turn telling, and commanding with himself. The child-robot baseline is similar to the child-trainer baseline but the robot was used as the second entity to participate in the turn taking, turn telling, and commanding evaluation. The two baselines were used to evaluate the SMD of the child-robot therapy since we wanted to compare the child-robot to the typical approach and to its own baselines.

The SMD results are shown in Table 6. For SC+DC. For SMD1, except ND+VD, all the effect sizes are more than 0.8 and it shows that the child had much improvement compared to the baseline. But in SMD2 and SMD3 there was no significant difference. Specially, in ND+VD the improvement is not considerable. It might be because of the simplicity of the task. Also in this item, the child had a little training for turn taking before our therapy. For RA+RAD the SMD2 shows large effect sizes for child improvement that is a little more in the child-robot case compared to child-trainer one.

It is also possible to directly compare the child-trainer with the child-robot interventions through SMD, rather than comparing them through the baselines. The results are shown in Table 7. For RA+RAD there is not good effect size and for ND+VD there is a medium effect size. But for SC+DC the effect size is very large.

In Table 8, the results for PEM are shown. As it can be seen for ND+VD, PEM is 0.47 which shows no difference to the typical approach, i.e. the child-trainer approach. However, RA+RAD is between 0.7 and 0.9 which shows moderate effective treatment compared to the child-trainer therapy. Finally, PEM for SC+DC is above 0.9 which shows that the child-robot therapy is highly effective treatment compared to the child-trainer approach.

Finally, the results for NAP are shown in Table 9. As it can be seen for ND+VD NAP is 0.52. For RA+RAD the result is better than ND+VD but not much better. Finally, SC+DC shows higher difference between the child-trainer and the child-robot therapies. As it can be seen in Figure. 2, RA+RAD for the child-robot therapy is at its maximum value, for most of the time, while it is changing for the child-trainer therapy.

### 3.2 Effect size analyses

To compare the proposed approach to the typical child-trainer therapy approaches, the effect sizes for ND+VD in turn-taking, RA+RAD in turn-telling, and SC+DC in commanding are calculated (Tables 4 to 9).

By evaluating the results, it can be seen that the child-robot therapy has been more effective than the child-trainer using the above analysis. Specially, for RA+RAD and SC+DC the differences are considerable in all effect sizes calculations in this study. For instance, child-robot therapy has very large SMD, PEM, and NAP effects in SC+DC sub-variable which shows very good effect in increasing these variables. But for ND+VD, the effect sizes were very small and not considerable.

Finally, despite the better numbers in child-robot commanding therapy compared to the child-trainer commanding therapy, the effect cannot be realistically measured since the baseline could not be measured due to the difficulty of the commanding task and the level of the subject.

It should be mentioned that the child was more cooperative in the child-robot therapy sessions compared to the child-trainer sessions. More specifically, there were child-trainer sessions in which the child was tired and barely paid attention to the trainer. However, he was still motivated to cooperate in the child-robot sessions.

### 3.3 Qualitative results

In this section, at first all interview forms are explained and then noticeable points extracted from videos are presented.

**The interview questions and the summary of the answers:**

– *Does the child learn better with the robot?*

*Answer:* The robot is useful as it provides sensory stimulus (auditory and visual).

— *Does the child learn faster with the robot?*

Answer: since the robot is attractive, children may pay more attention and learn faster compared to child-trainer therapy. However, it depends on the activity type and the purpose.

— *Can the robot be used by families at home for therapy purposes??*

Answer: The robot can be used in by well-educated families at homes. However, it should be noted that the robot should not replace the social interaction needed between family members.

— *What is your suggestion for increasing the efficacy of using the robot?*

*Answer:* adding more auditory stimulus and more motor functions for getting the attention of the children

— *Can the robot eliminate the need for therapists?*

*Answer: At the current stage, the robot cannot be used by its own. However, it is a very good complementary/assistive tool.*

**Points extracted from sessions' videos:** All sessions' recorded videos were reviewed and the subsequent points were found.

— The most important role of the robot, in the training sessions, was its motivating role in encouraging the child to participate in the sessions. In many sessions, it encouraged the child and caused his cooperation in the training. For example, in intervention session 8, the child was inattentive and defiant to trainer's instructions in the child-trainer scenario. In contrast, when he entered the child-robot scenario, he was so motivated that he even initiated interaction with the robot and spontaneously commanded (SC) the robot.

— The other important effect of the robot was that the child was using eye-contact with the trainer/robot in child-robot scenarios compared to the child-trainer scenarios.

— Finally, the child was more comfortable and better expressed his emotions in child-robot scenarios compared to the child-trainer scenario. For instance, after the robot correctly named a card in its turn, the child told to the robot "Bravo!" or "Well done to you!". It was interesting that the child learned these words of encouragement from the trainer/robot in the sessions and was able to use them correctly. Moreover, he got excited and more attentive when the robot said to him "Bravo!" than when the trainer said.

# 4 Conclusion and Future Work

In this paper, we presented our work on using RoboParrot to teach turn-taking skills to children with autism in order to investigate the effectiveness of the robot in therapy sessions. The child-robot therapy is designed based on a typical turn taking therapy in the form of a simple game between a child and a partner. Three variables, i.e. turn taking, turn telling, and commanding, were used to compare the child-robot therapy to a typical child-trainer therapy. The variables' effect sizes show improvement in both child-robot and child-therapist therapies. However, the effect sizes are larger in the child-robot therapy compared to the child-trainer therapy. Also, we prepared a questionnaire to evaluate the effectiveness of the robot from the therapist's point of view and its results show advantages of using the robot.

It should be mentioned that there were few sessions that the child did not cooperate with the therapist and was distracted. At the same session, when the child entered the child-robot therapy, he cooperated with the robot and the therapy session could be conducted. In other words, the robot was attractive enough to make the child attend the therapy session. It may indicate that the robot can be used as a complementary device, in therapy sessions, to make the therapy sessions more appealing to the children with autism.

For the future work, we need to test our approach on more children with autism in order to improve our approach and further validate it. In addition, the effect of the proposed therapy in real world situations, such as conversations or role play, should be evaluated. In other words, the proposed RAT protocol should be generalizable by children, especially children with autism. This may involve adding a variety of games and activities in order to better stimulate children's ability to generalize turn-taking skills. Finally, we would add more functionalities to the robot to extend the attractiveness of the robot.

**Acknowledgement**

We like to thank the Center for Treatment of Autism Disorder (CTAD) and its members for supporting this study. This project is funded by the grant number 123 by the Cognitive Sciences and Technologies council.

**Consent Form:**

<div dir="rtl">

بنام خدا

فرم رضایت آگاهانه برای تحقیقات کیفی

پیوست یک: فرم رضایت آگاهانه مشارکت کننده (اساتید ، بیماران ، دانشجویان)

همکار گرامی

این آزمون درراستای طرح پژوهشی اینجانب پگاه سلیمان دانشجوی دکتری رباتیک دانشگاه تهران انجام می گیرد. در این آزمون ارتباط و نحوه بازی کودک شما به منظور استفاده از ربات در درمان، مورد بررسی قرار خواهد گرفت.

مشارکت شما کاملا اختیاری می باشد در حین بازی تصویر کودک شما ضبط خواهد شد، ضبط تصویر فقط برای بررسی توسط متخصص، گردآوری داده ها و ثبت داده ها بوده و هیچ گونه استفاده دیگری از آنها نخواهد شد. این تصاویر در مرکز اتیسم نگهداری خواهد شد و اجازه خروج آن‌ها به هیچ عنوان داده نخواهد شد. شما می توانید از شرکت کودک خود در آزمون خودداری کنید و یا در زمان دلخواه آزمون را قطع نمایید. هیچ گونه هزینه ای جهت مشارکت در این پژوهش از شما دریافت نخواهد شد، هم چنین می توانید بخواهید در پایان، نتایج پژوهش برای شما ارسال شود.

پژوهشگر جهت انجام این پژوهش نیاز به یاری شما دارد و متعهد می گردد که در اجرای پژوهش، اخلاق پژوهش را حفظ نماید.

> شما می توانید اگر اشکال یا اعتراضی نسبت به دست اندرکاران یا روند پژوهش دارید با کمیته اخلاق در پژوهش دانشگاه علوم پزشکی ایران به آدرس : - تهران دانشگاه علوم پزشکی ایران ، بزرگراه شهید همت غرب بین تقاطع شیخ فضل اله و شهید چمران ستاد مرکزی طبقه 5 معاونت تحقیقات و فناوری تماس گرفته و مشکل خود را به صورت شفاهی یا کتبی مطرح نمایید.
>
> این فرم اطلاعات و رضایت آگاهانه در دو نسخه تهیه شده و پس از امضا یک نسخه در اختیار شما و نسخه دیگر در اختیار مجری قرار خواهد گرفت.

امضا مشارکت کننده..................................................

امضا پژوهشگر: پگاه سلیمان

</div>

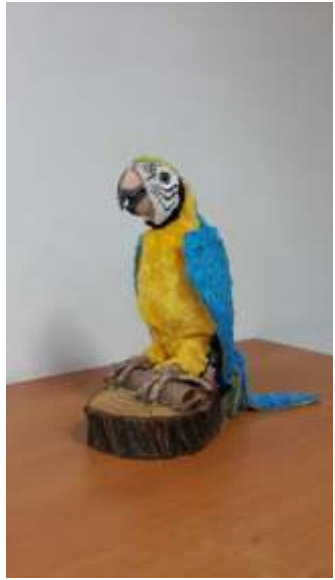

**Figure 1.** The RoboParrot

**Table 1.** Turn taking variables and sub-variables

| Variables | Sub-variables | Description |
|---|---|---|
| **Turn taking** | 1. Non-Directed (ND)<br>2. Directed | - Name body parts<br>- Name animals<br>- Name fruits |
| **Commanding** | 1. Spontaneous Command (SC)<br>2. Directed Command (DC)<br>3. No Command (NC) | - The subject commanded the partner to take his turn |
| **Turn telling** | 1. False Answer (FA)<br>2. Correct Answer (CA)<br>3. Correct Answer Directed (CAD): | - The subject tells the turn (your/my turn) |

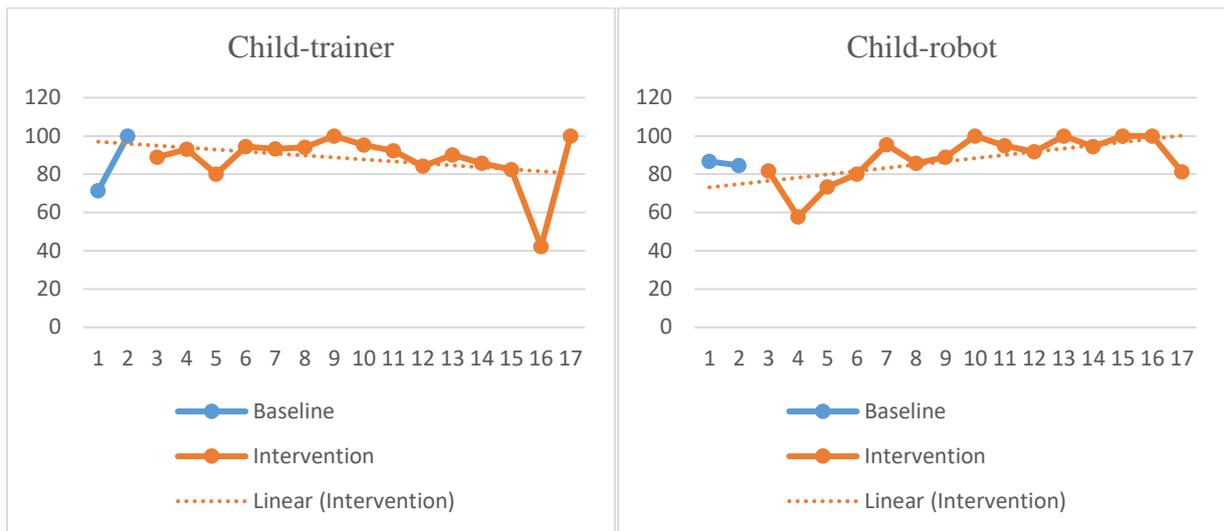

**Figure 2.** Chid-trainer and Child-robot Turn-taking ND+VD with the baselines

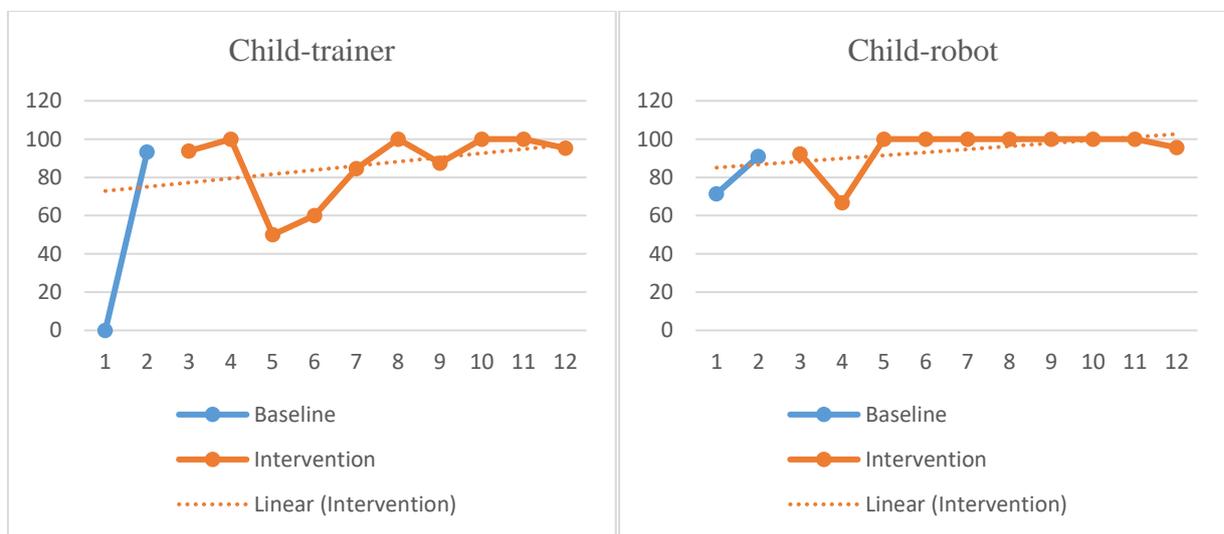

**Figure 3.** Chid-trainer and Child-robot Turn-taking RA+RAD with the baselines

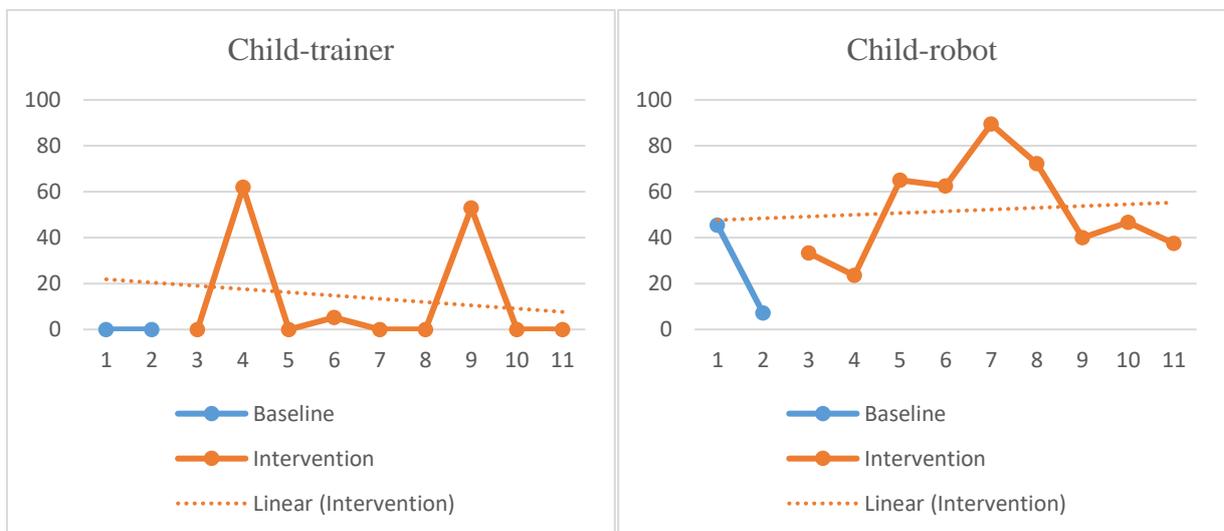

**Figure 4.** Chid-trainer and Child-robot Turn-taking SC+DC with the baselines

**Table 2.** The frequency of the indexes of turn taking, turn telling and command in child-trainer sessions

| Variables | Indices | Baselines | | | | | | | Intervention sessions | | | | | | | | | | |
|---|---|---|---|---|---|---|---|---|---|---|---|---|---|---|---|---|---|---|---|
| | | B 1 | B 2 | S 1 | S 2 | S 3 | S 4 | S 5 | S 6 | S 7 | S 8 | S 9 | S 10 | S 11 | S 12 | S 13 | S 14 | S 15 |
| turn taking | PD | 14.29 | 0.00 | 11.11 | 6.90 | 20.00 | 5.56 | 6.67 | 5.88 | 0.00 | 4.76 | 7.69 | 15.79 | 10.00 | 14.29 | 17.65 | 57.89 | 0.00 |
| | VD | 28.57 | 14.29 | 27.78 | 13.79 | 10.00 | 0.00 | 0.00 | 0.00 | 0.00 | 0.00 | 7.69 | 5.26 | 0.00 | 0.00 | 23.53 | 0.00 | 100.00 |
| | ND | 42.86 | 85.71 | 61.11 | 79.31 | 70.00 | 94.44 | 93.33 | 94.12 | 100.00 | 95.24 | 84.62 | 78.95 | 90.00 | 85.71 | 58.82 | 42.11 | 0.00 |

| Variables | Indices | Baselines | | | | | | | Intervention sessions | | | | | | | |
|---|---|---|---|---|---|---|---|---|---|---|---|---|---|---|---|---|
| | | B 1 | B 2 | B 3 | B 4 | B 5 | B 6 | S 1 | S 2 | S 3 | S 4 | S 5 | S 6 | S 7 | S 8 | S 9 | S 10 | S 11 |
| turn telling | FA | | | | | 100.00 | 6.67 | 6.25 | 0.00 | 50.00 | 40.00 | 15.38 | 0.00 | 12.50 | 0.00 | 0.00 | 4.76 | 0.00 |
| | RA | | | | | 0.00 | 73.33 | 93.75 | 58.82 | 50.00 | 20.00 | 46.15 | 40.00 | 62.50 | 50.00 | 33.33 | 66.66 | 0.00 |
| | RAD | | | | | 0.00 | 20.00 | 0.00 | 41.18 | 0.00 | 40.00 | 38.46 | 60.00 | 25.00 | 50.00 | 66.67 | 28.57 | 0.00 |

| Variables | Indices | Baselines | | | | | | | | Intervention sessions | | | | | | | | | |
|---|---|---|---|---|---|---|---|---|---|---|---|---|---|---|---|---|---|---|
| | | B 1 | B 2 | B 3 | B 4 | B 5 | B 6 | B 7 | B 8 | S 1 | S 2 | S 3 | S 4 | S 5 | S 6 | S 7 | S 8 | S 9 |
| command | SC | 0.00 | 0.00 | 0.00 | 0.00 | 0.00 | 0.00 | 0.00 | 0.00 | 0.00 | 47.62 | 0.00 | 0.00 | 0.00 | 0.00 | 0.00 | 0.00 | 0.00 |
| | DC | 0.00 | 0.00 | 0.00 | 0.00 | 0.00 | 0.00 | 0.00 | 0.00 | 0.00 | 14.29 | 0.00 | 5.26 | 0.00 | 0.00 | 52.94 | 0.00 | 0.00 |
| | NC | 100.00 | 100.00 | 100.00 | 100.00 | 100.00 | 100.00 | 100.00 | 100.00 | 100.00 | 38.09 | 100.00 | 94.74 | 100.00 | 100.00 | 47.06 | 100.00 | 100.00 |

Table 3. The frequency of the indexes of turn taking, turn telling and command in child-robot sessions

| Variables | Indices | Baselines | | | | | | | Intervention sessions | | | | | | | | | | | |
|---|---|---|---|---|---|---|---|---|---|---|---|---|---|---|---|---|---|---|---|---|
| | | B 1 | B 2 | S 1 | S 2 | S 3 | S 4 | S 5 | S 6 | S 7 | S 8 | S 9 | S 10 | S 11 | S 12 | S 13 | S 14 | S 15 | | |
| turn taking | PD | 13.33 | 7.69 | 18.18 | 42.31 | 26.67 | 20.00 | 4.55 | 14.29 | 11.11 | 0.00 | 5.00 | 8.33 | 0.00 | 5.56 | 0.00 | 0.00 | 18.75 | | |
| | VD | 66.67 | 84.62 | 63.64 | 50.00 | 6.67 | 20.00 | 0.00 | 0.00 | 16.67 | 0.00 | 0.00 | 0.00 | 0.00 | 11.11 | 0.00 | 6.67 | 56.25 | | |
| | ND | 20.00 | 0.00 | 18.18 | 7.69 | 66.67 | 60.00 | 95.45 | 85.71 | 72.22 | 100.00 | 95.00 | 91.67 | 100.00 | 83.33 | 100.00 | 93.33 | 25.00 | | |

| | | Baselines | | | | | | | Intervention sessions | | | | | | | |
|---|---|---|---|---|---|---|---|---|---|---|---|---|---|---|---|---|
| | | B 1 | B 2 | B 3 | B 4 | B 5 | B 6 | S 1 | S 2 | S 3 | S 4 | S 5 | S 6 | S 7 | S 8 | S 9 | S 10 | S 11 |
| turn telling | FA | | | | | 28.57 | 9.09 | 7.69 | 33.33 | 0.00 | 0.00 | 0.00 | 0.00 | 0.00 | 0.00 | 0.00 | 4.34 | 0.00 |
| | RA | | | | | 28.57 | 27.27 | 61.54 | 33.33 | 14.29 | 57.14 | 100.00 | 57.14 | 0.00 | 0.00 | 77.78 | 47.83 | 0.00 |
| | RAD | | | | | 42.86 | 63.64 | 30.77 | 33.33 | 85.71 | 42.86 | 0.00 | 42.86 | 100.00 | 100.00 | 22.22 | 47.83 | 100.00 |

| | | Baselines | | | | | | | | Intervention sessions | | | | | | | | | |
|---|---|---|---|---|---|---|---|---|---|---|---|---|---|---|---|---|---|---|---|
| | | B 1 | B 2 | B 3 | B 4 | B 5 | B 6 | B 7 | B 8 | S 1 | S 2 | S 3 | S 4 | S 5 | S 6 | S 7 | S 8 | S 9 | |
| command | SC | 0.00 | 0.00 | 0.00 | 0.00 | 0.00 | 0.00 | 0.00 | 0.00 | 0.00 | 5.88 | 30.00 | 50.00 | 36.84 | 66.67 | 26.67 | 13.33 | 18.75 | |
| | DC | 0.00 | 0.00 | 0.00 | 0.00 | 0.00 | 0.00 | 45.45 | 7.14 | 33.33 | 17.65 | 35.00 | 12.50 | 52.63 | 5.56 | 13.33 | 33.33 | 18.75 | |
| | NC | 100.00 | 100.00 | 100.00 | 100.00 | 100.00 | 100.00 | 54.55 | 92.86 | 66.67 | 76.47 | 35.00 | 37.5 | 10.53 | 27.77 | 60 | 53.34 | 62.5 | |

**Table 4.** Mean and Standard Deviation of both Therapies

| Variable groups | Sub-variable | Child-robot sessions | | | | Child-trainer sessions | | | |
| --- | --- | --- | --- | --- | --- | --- | --- | --- | --- |
| | | Baseline | | Intervention | | Baseline | | Intervention | |
| | | M | SD | M | SD | M | SD | M | SD |
| Turn taking | ND+VD | 85.65 | 1.45 | 88.35 | 12.01 | 85.72 | 20.20 | 87.72 | 13.93 |
| Turn telling | RA+RAD | 81.17 | 13.77 | 95.46 | 10.45 | 46.67 | 65.99 | 87.11 | 17.93 |
| Commanding | SC+DC | 26.30 | 27.09 | 52.24 | 21.31 | 0 | 0 | 13.35 | 25.15 |

**Table 5.** Three SMDs

| | Baseline | Intervention |
| --- | --- | --- |
| **SMD1** | Child-robot | Child-robot |
| **SMD2** | Child-trainer | Child-robot |
| **SMD3** | Child-trainer | Child-trainer |

**Table 6.** SMDs' effect sizes

| SMD | | | |
| --- | --- | --- | --- |
| Sub-variable | SMD1 | SMD2 | SMD3 |
| ND+VD | 0.32 | 0.16 | 0.12 |
| RA+RAD | 1.17 | 1.03 | 0.84 |
| SC+DC | 1.06 | 3.47 | 0.75 |

**Table 7.** SMD effect size-child-trainer as baseline and child-robot as intervention

| **Sub-variable** | **SMD'** |
| --- | --- |
| ND+VD | 0.05 |
| RA+RAD | 0.57 |
| SC+DC | 1.67 |

**Table 8.** PEM effect size

| **Sub-variable** | **PEM** |
| --- | --- |
| ND+VD | 0.47 |
| RA+RAD | 0.80 |
| SC+DC | 1.00 |

**Table 9.** NAP effect size

| **Sub-variable** | **NAP** |
| --- | --- |
| ND+VD | 0.52 |
| RA+RAD | 0.68 |
| SC+DC | 0.95 |